\documentclass{article}

\usepackage{microtype}
\usepackage{graphicx}
\usepackage{subfigure}
\usepackage{booktabs} 
\usepackage{xcolor}
\usepackage{arydshln}
 
\usepackage{hyperref}

\usepackage[accepted]{icml2020}

\usepackage[algo2e,ruled,linesnumbered,lined]{algorithm2e}

\usepackage[english]{babel}
\usepackage[utf8]{inputenc}
\usepackage{graphicx}
\usepackage{amssymb,todonotes}
\usepackage{wrapfig}
\usepackage{bm}
\usepackage{amsmath}

\DeclareMathOperator*{\argmin}{arg\,min}
\usepackage{float}
\usepackage{mathtools}

\newcommand\LC[1]{\textcolor{black}{#1}}

\icmltitlerunning{Online Learned Continual Compression with Adaptive Quantization Modules}

\begin{document}

\twocolumn[
\icmltitle{Online Learned Continual Compression with Adaptive Quantization Modules}

\icmlsetsymbol{equal}{*}

\begin{icmlauthorlist}
\icmlauthor{Lucas Caccia}{mcgill,mila,fb}
\icmlauthor{Eugene Belilovsky}{mila,udem}
\icmlauthor{Massimo Caccia}{mila,udem,eai}
\icmlauthor{Joelle Pineau}{mcgill,mila,fb}
\end{icmlauthorlist}

\icmlaffiliation{mcgill}{McGill}
\icmlaffiliation{udem}{University of Montreal}
\icmlaffiliation{mila}{Mila}
\icmlaffiliation{fb}{Facebook AI Research}
\icmlaffiliation{eai}{ElementAI}
\icmlcorrespondingauthor{Lucas Caccia}{lucas.page-caccia@mail.mcgill.ca}
\icmlcorrespondingauthor{Eugene Belilovsky}{eugene.belilovsky@umontreal.ca}

\icmlkeywords{Machine Learning, ICML}

\vskip 0.3in
]



\printAffiliationsAndNotice{}  

\begin{abstract}

We introduce and study the problem of Online Continual Compression, where one attempts to simultaneously learn to compress and store a representative dataset from a non i.i.d data stream, while only observing each sample once. 
A naive application of auto-encoders in this setting encounters a major challenge: representations derived from earlier encoder states must be usable by later decoder states.
We show how to use discrete auto-encoders to effectively address this challenge and introduce Adaptive Quantization Modules (AQM) to control variation in the compression ability of the module at any given stage of learning. This enables selecting an appropriate compression for incoming samples, while taking into account overall memory constraints and current progress of the learned compression. 
Unlike previous methods, our approach \textit{does not require any pretraining}, even on challenging datasets. We show that using AQM to replace standard episodic memory in continual learning settings leads to significant gains on continual learning benchmarks. Furthermore we demonstrate this approach with larger images, LiDAR, and reinforcement learning agents. 
\vspace{-10pt}
\end{abstract}

\section{Introduction}
\begin{figure*}
    \begin{center}
    \vspace{-10pt}
    \includegraphics[width=0.9\textwidth]{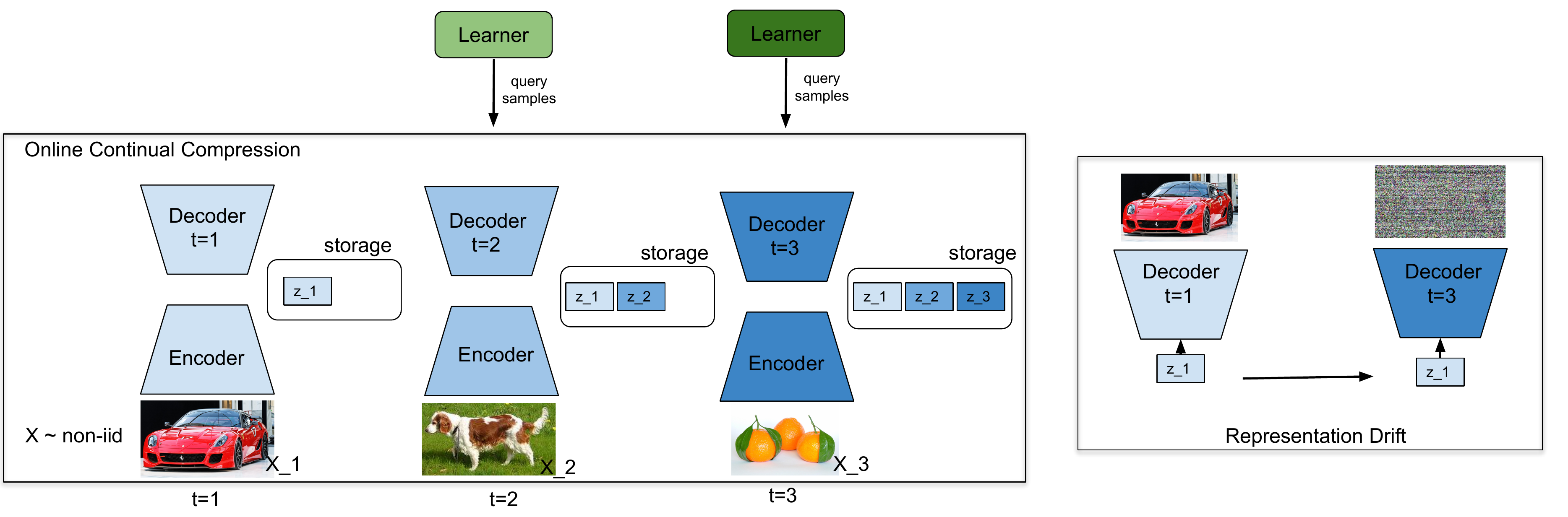}\vspace{-12pt}
    \caption{Illustration of the challenges in the Online Continual Compression problem. A model must be able to decode representations encoded by previous versions of the autoencoder, permitting anytime access to data for the learner. This must be accomplished while dealing with a time-varying data distribution and fixed memory constraints\vspace{-20pt}}
    \label{fig:online_continual_compression}
    \end{center}
\end{figure*}
Interest in machine learning in recent years has been fueled by the plethora of data being generated on a regular basis. Effectively storing and using this data is critical for many applications, especially those involving continual learning. In general, compression techniques can greatly improve data storage capacity, and, if done well, reduce the memory and compute usage in downstream machine learning tasks \citep{JPEG,Oyallon_2018_ECCV}. Thus, learned compression has become a topic of great interest \citep{theis2017lossy,balle2016end,johnston2018improved}. Yet its application in reducing the size of datasets bound for machine learning applications has been limited.

This work focuses on the following familiar setting: new training data arrives continuously for a learning algorithm to exploit, however this data might not be iid, and furthermore there is insufficient storage capacity to preserve all the data uncompressed. We may want to train classifiers, reinforcement learning policies, or other models continuously from this data as it's being collected, or use samples randomly drawn from it at a later point for a downstream task. For example, an autonomous vehicle (with bounded memory) collects large amounts of high-dimensional training data (video, 3D lidar) in a non-stationary environment (e.g. changing weather conditions), and overtime applies an ML algorithm to improve its behavior using this data. This data might be transferred at a later point for use in downstream learning. Current learned compression algorithms, e.g. \cite{torfason2018towards}, are not well designed to deal with this case, as their convergence speed is too slow to be usable in an online setting.

In the field of continual/lifelong learning \citep{thrun1995lifelong}, which has for now largely focused on classification, approaches based on storing memories for later use have emerged as some of the most effective in online settings \citep{lopez2017gradient,aljundi2018Online,chaudhry2018efficient,chaudhry2019continual,aljundi2019online}. These memories can be stored as is, or via a generative model \citep{shin2017continual}. Then, they can either be used for rehearsal \citep{chaudhry2019continual,aljundi2019online} or for constrained optimization \citep{lopez2017gradient,chaudhry2018efficient,aljundi2018Online}.
Indeed many continual learning applications would be greatly improved with replay approaches if one could afford to store all samples. These approaches are however inherently limited by the amount of data that can be stored

Learning a generative model to compress the previous data stream thus seems like an appealing idea. However, learning generative models, particularly in the online and non-stationary setting, continues to be challenging, and can greatly increase the complexity of the continual learning task. Furthermore, such models are susceptible to catastrophic forgetting \cite{aljundi2019online}. 
An alternate approach is to learn a compressed representation of the data, which can be more stable than learning generative models. 

 Learned compression in the online and non-stationary setting itself introduces several challenges illustrated in Fig~\ref{fig:online_continual_compression}. Firstly the learned compression module must be able to decode representations encoded by earlier versions of itself, introducing a problem we refer to as \textit{representation drift}. Secondly, the learned compressor is itself susceptible to catastrophic forgetting. Finally, the learned compression needs to be adaptive to maintain a prescribed level of reconstruction quality even if it has not fully adapted to the current distribution. 



In this work we demonstrate that the VQ-VAE framework \citep{van2017neural, razavi2019generating}, originally introduced in the context of generative modeling and density estimation, can be used online to effectively address representation drift while achieving high compression. Furthermore, when augmented with an internal replay mechanism it can overcome forgetting. Finally we use propose to use multiple gradient-isolated compression levels to allow the compressor to adaptively store samples at different compression scales, based on the amount of data, storage capacity, and effectiveness of the model in compressing samples. 


The main contributions of this work are: (a) we introduce and highlight the online learned continual compression (OCC) problem and its challenges. (b) We show how representation drift, one of the key challenges, can be tackled by effective use of codebooks in the VQ-VAE framework. (c) We propose an architecture using multiple VQ-VAE's, adaptive compression scheme, stream sampling scheme, and self-replay mechanism that work together to effectively tackle the OCC problem.  (d) We demonstrate this can yield state-of-the-art performance in standard online continual image classification benchmarks and demonstrate the applications of our OCC solution in a variety of other contexts. The code to reproduce our results is available at \url{https://github.com/pclucas14/adaptive-quantization-modules}

\vspace{-10pt}
\section{Related Work}
\vspace{-3pt}
\textbf{Learned compression} has been recently studied for the case of image compression. Works by \citet{theis2017lossy,balle2016end,johnston2018improved} have shown learned compressions can outperform standard algorithms like JPEG. These methods however are difficult to adapt for online settings as they do not directly address the challenges of the OCC problem (e.g. representation drift). 

\textbf{Continual Learning} research currently focuses on overcoming catastrophic forgetting (CF) in the supervised learning setting, with some limited work in the generative modeling and reinforcement learning settings. Most continual learning methods can be grouped into three major families. 

Some algorithms dynamically change the model’s architecture to incorporate learning from each task separately \citep{rusu2016progressive, li2018learning,fernando2017pathnet}. Although these methods can perform well in practice, their introduction of task-specific weights requires growing compute and memory costs which are problematic for the online setting. 
Another set of techniques employ regularization to constrain weights updates in the hope of maintaining knowledge from previous tasks. Notable methods in this class include \citep{kirkpatrick2017overcoming,huszar2017quadratic,zenke2017continual,nguyen2017variational, chaudhry2018riemannian}. However, this set of approaches has been shown to be inefficient in the online setting \cite{chaudhry2019continual}.

The last family of methods encapsulates all that have a mechanism to store information about the previous data distributions. This \textit{memory} then serves as a tool for the continual learner to rehearse previous tasks. The simplest instantiation of this method is to keep and sample from a buffer of old data to retrain the model after every update \citep{chaudhry2019continual}. This approach is widely used in RL where it is known as Experience Replay (ER) \citep{lin1993reinforcement, mnih2013playing, andrychowicz16nips}. Another method, known as Generative Replay (GR) \citep{shin2017continual}, uses generative modeling to store past task distributions. The continual learner then trains on generated samples to alleviate CF. Other notable examples are Gradient Episodic Memory (GEM) \citep{lopez2017gradient}, iCarl \citep{rebuffi2017icarl}, and Maximally Interfered Retrieval (MIR) \cite{aljundi2019online}, as well as \citep{aljundi2018Online, hu2018overcoming}. Most closely related to our work, \citet{scalable2017} consider compressing memories for use in the continual classification task. They also employ a discrete latent variable model but with the Gumbel approximation, which shows to be less effective than our approach. Furthermore a separate offline iid pre-training step for the learned compression is required in order to surpass the ER baseline, distinctly different from the online continual compression we consider. 

\textbf{Lidar compression} is considered in \cite{tu2019point} and \cite{caccia2018deep}. Both approaches use a similar projection from 3D $(x,y,z)$ coordinates to 2D cylindrical coordinates, and leverage deep generative models to compress the data. However, neither accounts for potential distribution shift, nor for online learning. In this work we show that using this 2D projection in conjunction with our model allows us to mitigate the two issues above for lidar data. 
\vspace{-10pt}

\section{Methodology}

In this section we outline our approach to the online continual compression problem. First we will review the VQ-VAE and highlight the properties making it effective for representational drift. Then we will describe our adaptive architecture, storage, and sampling scheme.

\subsection{Problem Setting: Online Continual Compression}
We consider the problem setting where a stream of samples $x \sim D_t$ arrives from different distributions $D_t$ changing over time $t=1 \dots T$. We have a fixed storage capacity of $C$ bytes where we would like to store the most representative information from all data distributions $D_1,...D_T$. There is notably a trade-off in quality of information versus the amount of samples stored. We propose to use a learned compression model, and most crucially, this  model must also be stored within the $C$ bytes, to encode and decode the data samples. 
Another critical requirement is that at anytime $t$ the content of the storage (data and/or compression model) be usable for downstream applications.
An important challenge, illustrated in Figure~\ref{fig:online_continual_compression}, is that the learned compression module will change over time, while we still need to be able to decode the memories in storage. 


\setlength{\textfloatsep}{9pt}
\begin{figure}
    \centering
    \vspace{-10pt}
    \includegraphics[width=0.9\linewidth]{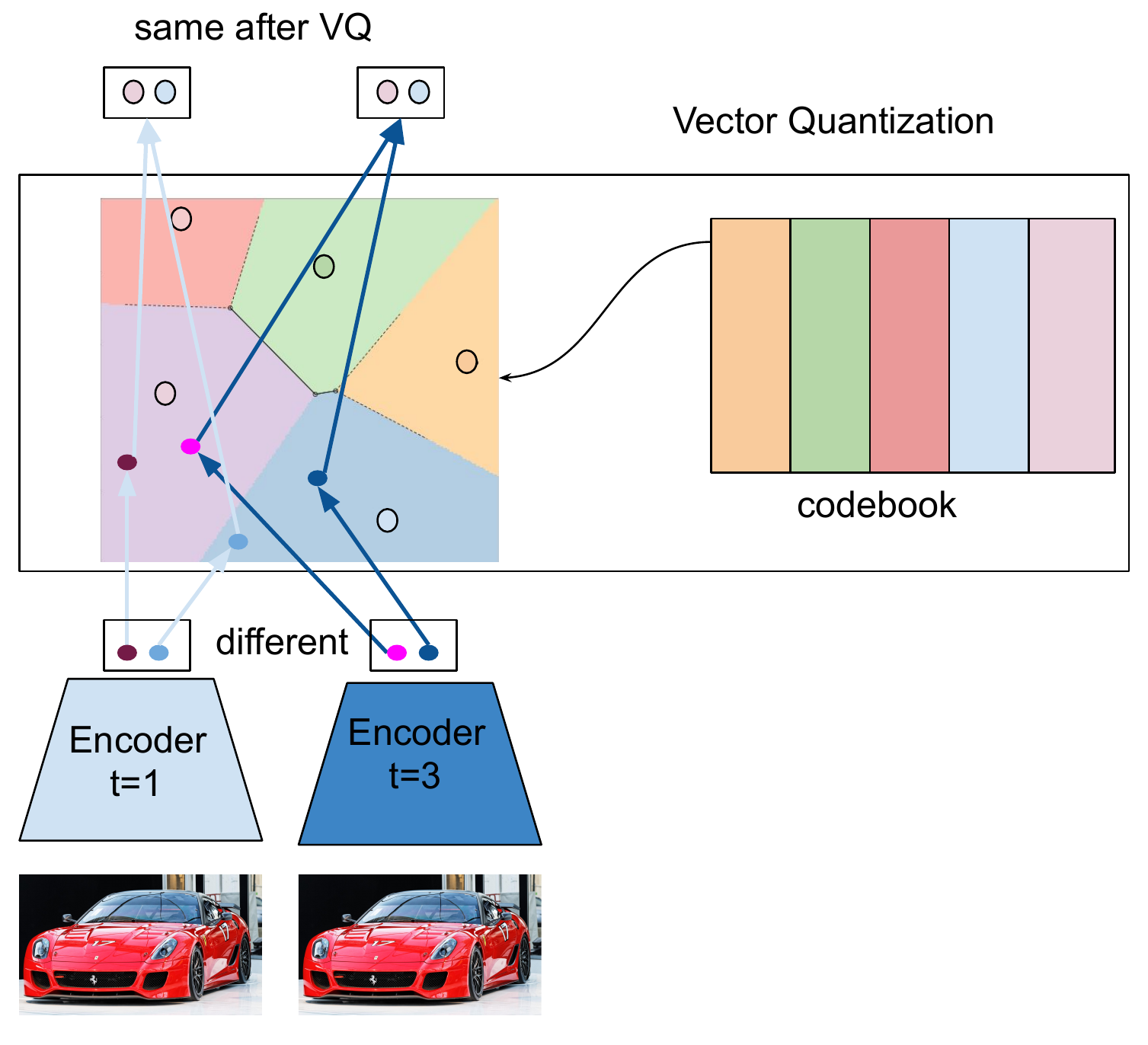}
    \vspace{-16pt}\caption{Illustration of reduced representation drift from Vector Quantization}
    \label{fig:vq_illustration}
\end{figure}

\subsection{Vector Quantized VAE for Online Compression}\label{sec:vqvae}
The VQ-VAE is a discrete auto-encoder which relies on a vector quantization step to obtain discrete latent representations. An embedding table, $E \in \mathbb{R}^{K \times D}$ consisting of $K$ vectors of size $D$, is used to quantize encoder outputs. Given an input (e.g. an RGB image), the encoder first encodes it as a $H_h \times W_h \times D$ tensor, where $H_h$ and $W_h$ denote the height and width of the latent representation. Then, every $D$ dimensional vector is quantized using a nearest-neighbor lookup on the embedding table. Specifically,  $z_{ij} = \argmin_{e \in E} || \texttt{enc}(x)_{ij} - e ||_2$, where $i,j$ refers to a spatial location. The output of the quantization step is then fed through the decoder. The gradient of this  non-differentiable step is approximated using the straight-through estimator. An important property to notice is that to reconstruct the input, only the $H_h \times W_h$ indices are required, thus yielding high compression \citep{van2017neural}.

Critically, the embedding tables are updated independently from the encoder and decoder, namely by minimizing $\min_{e} ||sg[\texttt{enc}(x)_{ij}] - e||$, where $sg$ is the stop gradient operator. 

Observe in the case of online compression, if the embedding table is fixed, then a change in the encoder parameters and therefore a change in the encoder output for a given input will not change the final quantized representation $z$, unless it is sufficiently large, thus we can observe that if the embeddings change slowly or are fixed we can greatly improve our control of the representational drift. This effect is illustrated in Figure~\ref{fig:vq_illustration}.
On the other hand we do need to adapt the embedding table, since randomly selected embeddings would not cover well the space of encoder outputs.



\begin{algorithm2e}\small
\SetAlgoLined
\DontPrintSemicolon
    \KwIn{Learning rate $\alpha$, \textsc{ExternalLearner} }
        \textbf{Initialize:} AQM Memory $\mathcal{M}$; AQM Parameters $\theta_{aqm}$ \\
        
        \vspace{1pt}
        
        \For{$t \in 1..T$}{
        
        \vspace{2pt}
        \,\,\,\texttt{\% Fetch data from current task}\\
        \For{$B_{inc}\sim D_t$}{
        \For{$n \in 1..N$}{
            $B \gets B_{inc}$ \\
            \If{$t > 1$}{
                
                \vspace{1pt}
                \,\,\,\texttt{\% Fetch data from buffer}\\
                $B_{re} \sim \textsc{Sample}(M,\theta_{aqm})$\\
                $B \gets (B_{inc}, B_{re})$ \\
            }
            \vspace{1pt}
            \texttt{\% Update AQM}\\
            $\theta_{aqm} \gets ADAM(\theta_{aqm}, B, \alpha)$ \\
            
            \vspace{1pt}
            {\color{blue} \texttt{\% Send data to external learner}}\\
            {\color{blue}$\textsc{ExternalLearner}(B)$}\\
            \lIf{$t > 1$}{
                \textsc{UpdateBufferRep}($M, \theta_{aqm}$) }
                
        \vspace{1pt}
        \texttt{\%Save current  indices}\\
        \textsc{AddToMemory}(${M}, B_{inc}, \theta_{aqm}$)\\
        }
    }
}
\caption{ \textsc{AQM Learning with Self-Replay}}
\label{algo:vqr}
\end{algorithm2e}

\subsection{Adaptive Quantization Modules}
To address issues of how to optimize storage and sampling in the context of Online Continual Compression we introduce Adaptive Quantization Modules (AQM). We use AQM to collectively describe the architecture, adaptive multi-level storage mechanism, and data sampling method used. Together they allow effectively constraining individual sample quality and memory usage while keeping in storage a faithful representation of the overall distribution.

AQM uses an architecture consisting of a sequence of VQ-VAEs, each with a buffer. 
The AQM approach to online continual compression is overall summarized in Algorithm ~\ref{algo:vqr} (note $\textsc{AddToMemory}$ is described in Appendix~\ref{sec:mem}). Incoming data is added to storage using an adaptive compression scheme described in Algorithm~\ref{algo:adapcomp}. Incoming data is also used along with randomly sampled data from storage (self-replay) to update the current AQM model. The randomly sampled data also updates its representation in storage as per Algorithm ~\ref{algo:update}. As illustrated by the optional lines in blue, Algorithm ~\ref{algo:vqr} can run concurrently with a downstream learning task (e.g. online continual classification) which would use the same batch order. It can also be run independently as part of a data collection. In the sequel we give further details on all these elements  

\subsubsection{Architecture and Training}
\begin{figure}
    \begin{center}
    \includegraphics[width=0.35\textwidth]{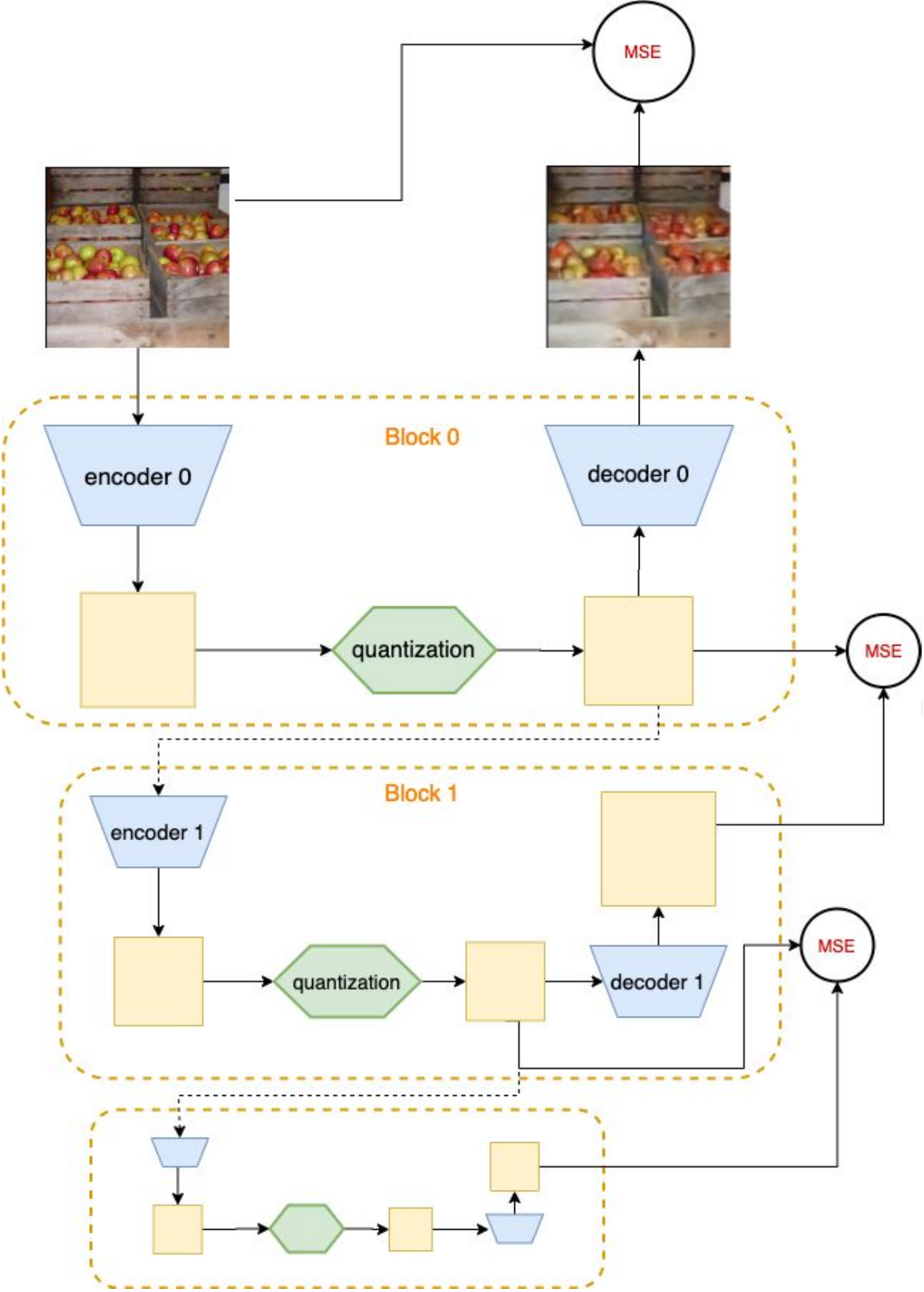}
    \caption{Architecture of Adaptive Quantization Modules. Each level uses its own loss and maintains its own replay buffer. Yello dotted lines indicate gradient isolation between modules}
    \label{fig:soft_modules}
    \end{center}
\end{figure}

Each AQM module contains a VQ-VAE and a corresponding buffer of adaptive capacity. A diagram of the architecture is given in Figure~\ref{fig:soft_modules}. We will denote the output after quantization of each module $i$ as $z_q^i$ and the set of codebook indexes used to obtain $z_q^i$ as $a^i$. Note that $a^i$ are the discrete representations we actually store. Each subsequent module produces and stores an $a^i$ requiring fewer bits to represent.

For RGB images, the compression rate at a given level is given by
$
    \frac{H \times W \times 3 \times  \log_2 (256) }
         {N_c \times {H_h}_i \times {W_h}_i \times \left \lceil  {\log_2 {(K_i)}} \right \rceil}
$. 
Here $K_i$ is the number of embeddings in the codebooks, ($H_{hi}$, $W_{hi}$) the spatial dimension of the latent representation and ${N_c}_i$ the number of codebooks.

VQVAE-2 \citep{razavi2019generating} also uses a multi-scale hierarchical organization, where unlike our AQM, the top level models global information such as shape, while the bottom level, conditioned on the top one, models local information. While this architecture is tailored for generative modeling, it is less attractive for compression, as both the bottom and top quantized representations must be stored for high quality reconstructions. Furthermore in AQM each module is learned in a greedy manner using the current estimate of $z_q^{(i-1)}$ without passing gradients between modules similar to \cite{belilovsky2019decoupled,nokland2019training}. A subsequent module is not required to build representations which account for all levels of compression, thus minimizing interference across resolutions.  This allows the modules to each converge as quickly as possible with minimal drift at their respective resolution, particularly important in the online continual learning case.

\setlength{\textfloatsep}{0pt}
\setlength{\floatsep}{1pt}
\begin{algorithm2e}[t]\small

\SetAlgoLined
  \DontPrintSemicolon
    \KwIn{datapoint $x$, AQM with $L$ modules, threshold $d_{th}$} 
 \texttt{\% Forward all modules, store encodings} \\
$\{z_q^i, a^i\}_{i=1..L}$= \textsc{Encode}$(x)$ \\
\For {$i \in L...1$}{
  \texttt{\% Decode from level i to output space} \\
   $\hat{x} = \textsc{Decode}(z_q^i)$\\
  \texttt{\% Check reconstruction error} \\
    \lIf{$\textsc{MSE}(\hat{x}, x) < d_{th}$}{
        $\textbf{return } a_i, \ i$
    }
}
\texttt{\% Otherwise, return original input} \\
\textbf{return} $x, 0$
\caption{\textsc{AdaptiveCompress}}
\label{algo:adapcomp}
\end{algorithm2e}

\begin{algorithm2e}[t] \small
\SetAlgoLined
  \DontPrintSemicolon
    \KwIn{Memory $\mathcal{M}$, AQM with $L$ levels, data $D$, distortion threshold $d_{th}$}
\For{$x \in D$}{
    $hid_x$, $block_{id}$ = \textsc{AdaptiveCompress}($x$, AQM, $d_{th}$) \\
    
    \texttt{\% Delete Old Repr.} \\
    \textsc{Delete}($\mathcal{M}$[$x$]) \\
    \texttt{\% Add new one} \\
    \textsc{Add}($\mathcal{M}$, $hid_x$) 
}
 \caption{UpdateBufferRep}
\label{algo:update}
    
\end{algorithm2e}
\setlength{\textfloatsep}{1pt}


\subsubsection{Multi-Level Storage}\label{sec:adap}
Our aim is to store the maximum number of samples in an allotted $C$ bytes of storage, while assuring their quality, and our ability to reconstruct them. 
Samples are thus stored at different levels based on the compressors' current ability. 
The process is summarized in Algorithm ~\ref{algo:adapcomp}. 

 Such an approach is particularly helpful in the online non-stationary setting, allowing knowledge retention before the compressor network has learned well the current distribution. Note in Alg.~\ref{algo:adapcomp} samples can be completely uncompressed until the first module is able to effectively encode them.  This can be crucial in some cases, if the compressor has not yet converged, to avoid storing poorly compressed representations.  Further taking into account that compression difficulty is not the same for all datapoints, this allows use of more capacity for harder data, and fewer for easier.

 We also note, since we maintain stored samples at each module and the modules are decoupled, that such an approach allows to easily distribute training in an asynchronous manner as per \citet{belilovsky2019decoupled}.
\subsubsection{Self-Replay and Stream Sampling}
As shown in Alg.~\ref{algo:vqr} our AQM is equipped with an internal experience replay mechanism \cite{mnih2013playing}, which reconstructs a random sample from storage and uses it to perform an update to the AQM modules, while simultaneously freeing up overall memory if the sample can now be compressed at a later AQM module. This has the effect of both reducing forgetting and freeing memory. In practice we replay at the same rate as incoming samples arrive and thus replay will not increase asymptotic complexity of the online learning. Finally, for efficiency the replay can be coupled to an external online learner querying for random samples from the overall memory.

Since we would like the AQM to work in cases of a fixed memory capacity it must also be equipped with a mechanism for selecting which samples from the stream to store and which to delete from memory. Reservoir Sampling (RS) is a simple yet powerful approach to this problem, used successful in continual learning \cite{chaudhry2019continual}. It adds a sample from the stream with prob. $p=\frac{\text{buffer capacity}}{\text{points seen so far}}$ while remove a random sample. However, RS is not directly compatible with AQM primarily because the amount of samples that can be stored varies over time. This is because samples at different levels have different memory usage and memory can be freed by replay. We thus propose an alternative scheme, which maximally fills available memory and selects non-uniformly samples for deletion. Specifically when a larger amount of samples are added at one point in the stream, they become more likely to be removed. The details of this stream sampling method are provided in Appendix~\ref{sec:mem}.

\subsubsection{Drift Control via Codebook Stabilization}
As mentioned previously, a good online compressor must control its representational drift, which occurs when updates in the auto-encoder parameters creates a mismatch with the static representations in the buffer. Throughout the paper we measure representational drift by comparing the following time varying quantity: $\textsc{Drift}_t(z) = \textsc{Recon Err} \big(\texttt{Decode}(\theta_t; z), x\big)$
where $\theta_t$ the model parameters at time $t$ and $(z, x)$ is a stored compressed representation and its original uncompressed datapoint respectively. For all experiments on images, \textsc{Recon Err} is simply the mean squared error. 

As illustrated in Sec \ref{sec:vqvae} a slow changing codebook can allow to control drifting representations. This can be in part accomplished by updating the codebook with an exponential moving average as described in \citep[Appendix A]{van2017neural}, where it was used to reduce the variance of codebook updates. This alone is insufficient to fully control drift, thus once a given module yields satisfactory compressions on the data stream, we freeze the module's embedding matrix but leave encoder and decoder parameters free to change and adapt to new data. Moreover, we note that fixing the codebook for a given module does not affect the reconstruction performance of subsequent modules, as they only need access to the current module's decoder which can still freely change. 
\vspace{-10pt}
\section{Experiments}

We evaluate the efficacy of the proposed methods on a suite of canonical and new experiments. In  Section \ref{sec:can_cl} we present results on standard supervised continual learning benchmarks on CIFAR-10. In Section~\ref{sec:off} we evaluate other downstream tasks such as standard iid training applied on the storage at the end of online continual compression. For this evaluation we consider larger images from Imagenet, as well as on lidar data. Finally we apply AQM on observations of an agent in an RL environment. 

\subsection{Online Continual Classification}


\begin{table*}
    \centering
    \resizebox{0.8\textwidth}{!}{%
    \begin{tabular}{ll}
    \begin{tabular}{c c c}
&\multicolumn{2}{c}{Accuracy ($\uparrow$)} \\\hline
                 & $M=20$ & $M=50$  \\\hline \hline
    iid online     &$60.8\pm1.0$ &$60.8\pm1.0$\\
    
    iid offline     &$79.2\pm 0.4$ &$79.2\pm 0.4$\\\noalign{\hrule height 1pt}
    
    GEM   \citep{lopez2017gradient}   &$16.8 \pm 1.1$& $17.1 \pm 1.0$\\
    
    iCarl (5 iter)  \citep{rebuffi2017icarl}   &$28.6 \pm 1.2$ & $33.7 \pm 1.6$ \\
    
    fine-tuning &$18.4\pm0.3$& $18.4\pm0.3$\\
    
    ER&$27.5\pm 1.2$ &$33.1\pm 1.7$\\ 
     
    ER-MIR \citep{aljundi2019online}& $29.8\pm1.1$ &$40.0 \pm 1.1$ \\  \hdashline  
    
    ER-JPEG & $33.9\pm1.0$ & $43.1\pm0.6$ \\
    
    Gumbel AE \citep{riemer2018learning} & $25.5\pm2.0$ &$28.8 \pm 2.9$ \\
   
    AQM (ours) & $\bm{43.5 \pm 0.7}$ & $\bm{47.0 \pm 0.8}$ \\ 
\end{tabular}
&
\begin{tabular}{c c}
\multicolumn{2}{c}{Forgetting ($\downarrow$)} \\\hline
    $M=20$ & $M=50$  \\\hline \hline
    
    N/A &N/A\\
    N/A&N/A\\\noalign{\hrule height 1pt}
    
    $73.5 \pm 1.7$ & $70.7 \pm 4.5$ \\
    
    $49 \pm2.4$& $40.6\pm1.1$\\
    
    $85.4 \pm 0.7$  & $85.4 \pm 0.7$\\
    
    $50.5\pm 2.4$&$35.4\pm 2.0$ \\ 
    
    $50.2\pm 2.0$& $30.2 \pm 2.3$ \\ \hdashline
    
    $54.8\pm1.2$ & $44.3 \pm 0.9 $ \\
    
    $71.5\pm 2.8$&$ 67.2 \pm 3.9$ \\
    
    $\bm{23.0 \pm 1.0} $ & $\bm{19.0 \pm 1.4}$ \\

\end{tabular}
\end{tabular}
} 
\caption{Shared head results on disjoint CIFAR-10. Total memory per class $M$ measured in sample memory size. We report  (a) Accuracy, (b) Forgetting (lower is better). }\vspace{-18pt}
\label{tab:cifar10_er_main}
\end{table*}

\label{sec:can_cl}
Although CL has been studied in generative modeling \citep{ramapuram2017lifelong,lesort2018generative,Zhai2019LifelongGC,lesort2019marginal} and reinforcement learning \citep{kirkpatrick2017overcoming,fernando2017pathnet,riemer2018learning}, supervised learning is still the standard for evaluation of new methods. Thus, we focus on the online continual classification of images for which our approach can provide a complement to experience replay. In this setting, a new task consists of new image classes that the classifier must learn, while not forgetting the previous ones. The model is only allowed one pass through the data \citep{lopez2017gradient,chaudhry2018efficient,aljundi2019online,chaudhry2019continual}. The online compression here takes the role of replay buffer in replay based methods such as \cite{chaudhry2019continual,aljundi2019online}. We thus run Algorithm~\ref{algo:vqr}, with an additional online classifier being updated performed at line 15.

Here we consider the more challenging continual classification setting often referred to as using a \textit{shared-head} \citep{aljundi2019online,farquhar2018towards,aljundi2018Online}. Here the model is not informed of the task (and thereby the subset of classes within it) at test time. This is in contrast to other (less realistic) CL classification scenarios where the task, and therefore subset of classes, is provided explicitly to the learner \citep{farquhar2018towards,aljundi2019online}.


For this set of experiments, we report accuracy, i.e. $\frac{1}{T} \sum_{i=1}^T R_{T,i}$, and forgetting, i.e. $\frac{1}{T-1} \sum_{i=1}^{T-1} \max(R_{:,i}) - R_{T,i}$
with $R \in \mathbb{R}^{T \times T}$ representing the accuracy matrix where $R_{i,j}$ is the test classification accuracy on task $j$ when task $i$ is completed.

\vspace{-8pt}
\paragraph{Baselines}
A basic baseline for continual supervised learning is Experience Replay (\textbf{ER}). It consists of storing old data in a buffer to replay old memories. Although relatively simple recent research has shown it is a critical baseline to consider, and in some settings is actually state-of-the-art \citep{chaudhry2019continual,aljundi2019online,rolnick2018experience}. AQM can be used as an add-on to ER that incorporates online continual compression. \LC{We also compare against ER with standard JPEG compression}.
In addition we consider the following baselines. \textbf{iid online} (upper-bound)
 trains the model with a single-pass through the data on the same set of samples, but sampled iid. \textbf{iid offline} (upper-bound) evaluates the model using multiple passes through the data, sampled iid. We use 5 epochs in all the experiments for this baseline. \textbf{fine-tuning} trains continuously upon arrival of new tasks without any forgetting avoidance strategy. \textbf{iCarl} \citep{rebuffi2017icarl} incrementally classifies using a nearest neighbor algorithm, and prevents catastrophic forgetting by using stored samples. \textbf{GEM} \citep{lopez2017gradient} uses stored samples to avoid increasing the loss on previous task through constrained optimization. It has been shown to be a strong baseline in the online setting. It gives similar results to the recent A-GEM \cite{chaudhry2018efficient}. \textbf{ER-MIR} \citep{aljundi2019online} controls the sampling of the replays to bias sampling towards samples that will be forgotten. We note that the ER-MIR critera is orthogonal to AQM, and both can be applied jointly. \textbf{Gumbel AE} \cite{riemer2018learning} learns an autoencoder for ER using the Gumbel softmax to obtain discrete representations. 
 
 
 We evaluate with the standard CIFAR-10 split \citep{aljundi2018Online}, where 5 tasks are presented sequentially, each adding two new classes. Evaluations are shown in Table~\ref{tab:cifar10_er_main}. 
Due to our improved storage of previous data, we observe significant improvement over other baselines at various memory sizes. 
We can contrast AQM's performance with ER's to understand the net impact of our compression scheme. Specifically, AQM improves over ER by 16.0\% and 13.9\% in the M=20 and M=50 case, highlighting the effectiveness of online compression. \LC{Our approach is also superior in forgetting by a significant margin in both memory settings.}

To compare directly to reporting in \cite{riemer2018learning} we also benchmarked our implementation on the Incremental CIFAR-100 multi-head experiment \cite{lopez2017gradient} with the same settings as in \cite{riemer2018learning}. By using AQM we were able to get \textbf{65.3} vs the reported \textbf{43.7} using a buffer of size 200. To specifically isolate the advantage of gumbel softmax versus the vector quantization for drift, we replaced the vector quantization approach with gumbel softmax in an AQM. We observed signficantly less drift in the case where vector quantization is used. Full details of this experiment are described in the supplementary materials along with visualizations.



The CIFAR-10 dataset has a low resolution ($3 \times 32 \times 32$) and uses a lot of data per task (10K samples). These two characteristics might leave the online compression problem easier than in a real-life scenario. Specifically, if the first tasks are long enough and the compression rate is not too large, the model can quickly converge and thus not incur too much representation drift. Indeed, we found that using a single module is already sufficient for this task. For these reasons, we now study the AQM in more challenging settings presented in the next section.

\subsection{Offline Evaluation on Larger Images} \label{sec:off}

\setlength{\textfloatsep}{9pt}
\begin{table}
    \centering
    \resizebox{0.85\linewidth}{!}{%
    \begin{tabular}{c|c}
&Accuracy \\\hline \hline
    RS &$5.2 \pm 0.2$\\
    2 Module AQM (ours) & $\bm{23.2 \pm 1.1}$ \\\hdashline
    Ablate 2nd Module & $20.5 \pm 1.3$ \\
    Ablate Fixing Codebook & $19.2 \pm 0.6$  \\ 
    Ablate Decoupled Training & $16.5 \pm 0.7$ \\
    Ablate Adaptive Compression  & $13.1 \pm 3.2$ \\\hline

\end{tabular}
}
\caption{Imagenet offline training evaluation from online continual compression. We see a clear gain over a standard Reservoir sampling approach. We then ablate each component of our proposal showing each component is important. Note storage used in each experiment is identical (including accounting for model sizes).}\vspace{2pt}
\label{tab:offline}
\end{table}

Besides the standard continual classification setup, we propose several other evaluations to determine the effectiveness of the stored data and compression module after learning online compression. We also perform a detailed ablation to study the efficacy of each component in AQM.
\paragraph{Offline training on Imagenet} We compare the effectiveness of the stored memories of AQM after a certain amount of online continual compression. We do this by training in a standard iid way an offline classification model using only reconstructions obtained from the storage sampled after online continual compression has progressed for a period of time.  In each case we would have the same sized storage available. We note that simply having more stored memories does not amount to better performance as their quality may be severely degraded and affected by drift.

Using this evaluation we first compare a standard reservoir sampling approach on uncompressed data to a 2 module AQM using the same size storage. We observe that performance is drastically increased using the compressed samples. We then use this to perform a series of ablations to demonstrate each component of our proposal is important. Specifically (a) we restrict AQM to have only one module, (b) instead of decoupled training we train modules end-to-end,  (c) we remove adaptive compression, thus all samples are stored in the most compressed block, regardless of quality, and (d) we do not stabilize the codebook, the embedding matrices of every block are never fixed. We observe that all these elements contribute to successfully storing a representative set of data for the distribution online.

\setlength{\textfloatsep}{9pt}
\begin{figure}
    \centering
    \includegraphics[width=0.45\textwidth]{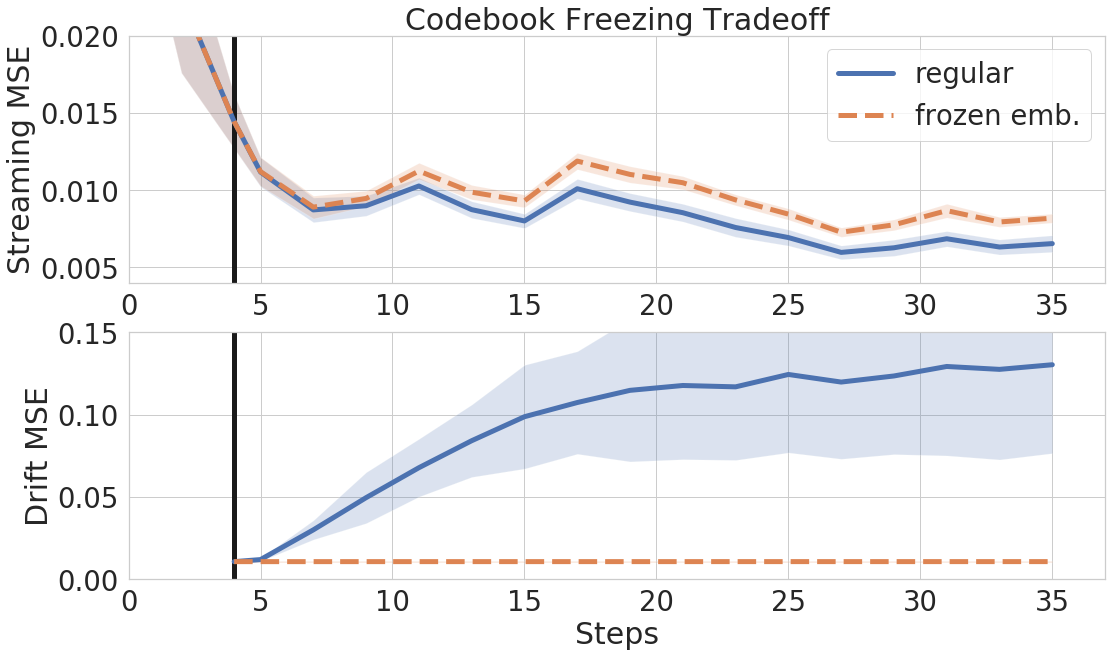}
    \vspace{-10pt}
    \caption{Impact of codebook freezing. Vertical black line indicates freezing point. We see that AQM is still able to adapt and reduce its reconstruction loss, while having stable compressed representations. Results averaged over 5 runs}
    \label{fig:ablate_2}
\end{figure}

\vspace{10pt}

\paragraph{Drift Ablation} We have seen the importance of codebook freezing when dealing with high dimensional datasets. However, judging solely from the final downstream task performance it's difficult to see if the model continues adapting after freezing. As alluded in Sec~\ref{sec:vqvae} there is a tradeoff between keeping recoverable representations and a model's ability to continue to adapt. To shed some light on this, we run the following experiment: we run a vanilla VQ-VAE on the same 20 task mini-imagenet stream, without storing any samples. When it reaches a pre-specified performance threshold, we fix the codebook, and store compressed \textit{held-out} data from the first task. We then continue to update the VQ-VAE parameters, and the memory is kept fixed for the rest of the stream. We apply self-replay but no other AQM mechanisms (e.g. no sampling from the input stream and no adaptive compression). 

We monitor how well the VQ-VAE can adapt by looking at the streaming reconstruction cost, measured on the incoming data before an update. We also monitor the drift of samples stored in the buffer. Results are presented in Figure \ref{fig:ablate_2}. They demonstrate that drift is controlled by stabilizing the codebook, while the model can still improve at nearly the same rate. Further analysis, \LC{along with an additional experiment showcasing the robustness of vector quantization to small perturbations} is included in the Appendix.

\paragraph{LiDAR} \LC{Range data enables autonomous vehicles to scan the topography of their surrounding, giving precise measurements of an obstacle's relative location. In its raw form, range data can be very large, making it costly to transmit in real time, or for long term storage. Equipping self-driving cars with a good lidar compressor can enable fast vehicle-to-vehicle (V2V) communication, 
leading to safer driving \cite{eckelmann2017v2v}. Moreover, since data collected by autonomous vehicles can be highly non-stationary (new objects on the road, changing weather or traffic conditions), having a compressor which can quickly adapt to this distribution change will reduce the required memory for storage (or bandwidth for real time transmission).}

\setlength{\textfloatsep}{9pt}
\begin{figure}[t]
        \centering
        \includegraphics[width=.4\textwidth]{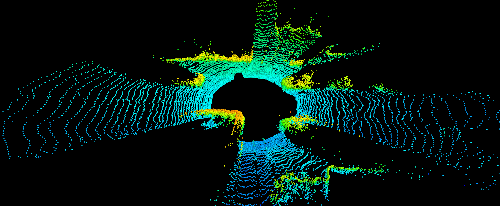} \ \ \ 
        \includegraphics[width=.4\textwidth]{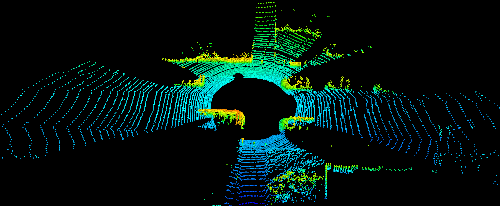} \vskip 2mm

\caption{Top: Sample decoded from the buffer at the end of training \LC{from scratch} (32x compression rate). Bottom: Original lidar} 
\label{fig:lidar_qualitative}
\end{figure}

\LC{We proceed to train AQM on the Kitti Dataset \citep{geiger2013vision}, which contains 61 LiDAR scan recordings, each belonging to either the ``residential", ``road", ``city" environments.} The data is processed as in \cite{caccia2018deep}, where points from the same elevation angle are sorted in increasing order of azimuth angle along the same row. This yield a 2D grid, making it compatible with the same architecture used in the previous experiments. As in \citep{caccia2018deep, tu2019point}, we report the reconstruction cost in Symmetric Nearest Neighbor Root Mean Squared Error (SNNRMSE) which allows to compare two point clouds. Note AQM can also be adapted to use task relevant criteria besides $\texttt{MSE}$. 

\LC{We consider two settings. In the first, we train AQM \textit{from scratch} on a data stream consisting of recordings from all three environments. We present (once) all the recordings of an environment before moving on to another, in order to maximise the distribution shift.} We show qualitative results in Figure~\ref{fig:lidar_qualitative} and in the supplementary materials. Observe that we are able to effectively reconstruct the LiDAR samples and can easily tradeoff quality with compression. Overall we obtain \textbf{18.8 cm} SNNRMSE with \LC{$32 \times$ compression}, which lies in a range that has been shown in \citep{tu2019point} to be sufficient to enable SLAM localization with very minimal error.

\setlength{\textfloatsep}{9pt}
\begin{table}
    \centering
    \resizebox{0.75\linewidth}{!}{%
    \begin{tabular}{c|c}
&Size in $\texttt{Mb}$ \\\hline \hline
    Raw & $1326.8$\\
    Gzip & $823.0$ \\ \hdashline
    AQM & $35.5 \pm .06$ \\
    AQM + finetune & $ 33.0 \pm .07$  \\
    AQM + finetune + PNG & $27.9 \pm .01$ \\\hline

\end{tabular}
}
\caption{Compression results for the data transmission of the city lidar recordings. We require that each compressed scan has an SNNRMSE under 15 cm.}
\label{tab:lidar_online}
\end{table}

\LC{In the second setting, we wish to simulate a scenario where some data is available a priori for the model to leverage. However, this data is limited and does not cover all the possible modalities to which an autonomous vehicle could be exposed. To this end, we pretrain AQM in a fully offline iid manner on the road and residential recordings. We then simulate the deployment of the compressor on a vehicle, where it must compress and transmit in real time the lidar data feed from a new distribution. We therefore stream the held-out city recordings and show that AQM can be fine-tuned on the fly to reduce the required bandwidth for data transmission. Quantitative results are presented in table \ref{tab:lidar_online}. We ensure that the reconstructed lidar scans have a SNNRMSE smaller than 15.0 cm.
Moreover, since the stored representations in AQM are 2D and discrete, we can apply lossless compression schemes such as Portable Network Graphics (PNG).}

\subsection{Atari RL Environments}
\begin{figure}
    \centering
    \vspace{-4pt}
    \includegraphics[width=0.5\linewidth]{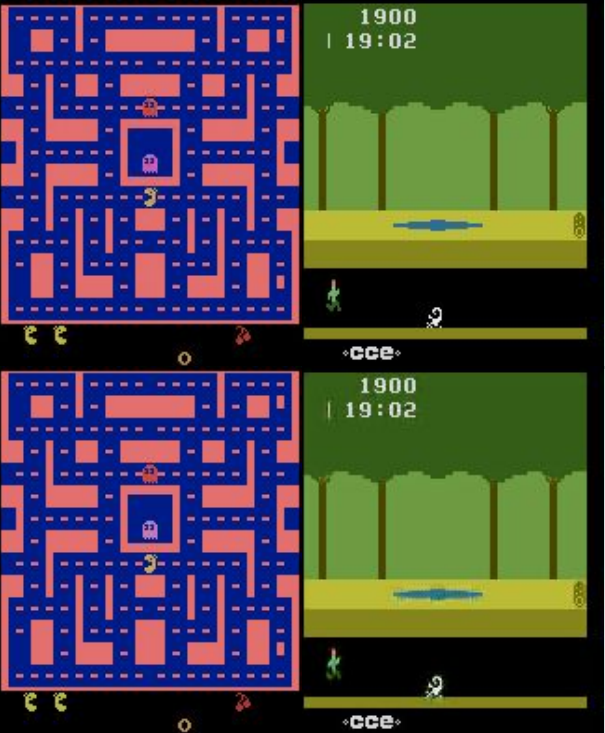}\vspace{-8pt}
    \caption{Top: original. Bottom: reconstructed from AQM}\vspace{3pt}
    \label{fig:atari}
\end{figure}

Another application of online continual compression  is for preserving the states of an reinforcement learning agent operating online. These agents may often learn new tasks or enter new rooms thus the observations will often be highly non-iid. Furthermore many existing reinforcement learning algorithms already rely on potentially large replay buffers which can be prohibitive \cite{dqn2014mnih,rolnick2018experience} to run and may greatly benefit from an approach such as the AQM to run concurrently with reinforcement learning algorithms. We thus perform a proof of concept for the AQM for storing the state sequence encountered by an RL learner in the atari environment\cite{bellemare2013arcade}. We use the dataset and tasks introduced in \cite{anand2019unsupervised}, which runs a random or learned policy in the atari environments and provides a set of classification tasks to evaluate whether key information about the state is preserved. Results are shown Table~\ref{tab:rl}. We run the online learning with the AQM on the data stream observed by the random agent. We use the same observations and optimization as in \cite{anand2019unsupervised} and report the F1 results of a linear probe directly on states for our reconstructions after online compression and the originals. Results for 3 environments are shown in Table~\ref{tab:rl} and examples in in Fig~\ref{fig:atari} and the Appendix. We find that AQM can well preserve the critical information while compressing the state by 16x. The reference accuracies achieved by our classifier are similar to those in \cite{anand2019unsupervised}. However, we do not control for the representation size unlike those evaluations of various unsupervised models. 
\begin{table}
    \centering
    \resizebox{0.68\linewidth}{!}{%
    \begin{tabular}{c|c|c} 
        Game & Cls Input & F1  \\\hline \hline 
        Pong & Orig. State & 86.7 \\
                & AQM Recon & 86.8 \\\hdashline
        Ms Pacman & Orig. State  & 89.4 \\
                & AQM Recon & 88.3 \\\hdashline
        Pitfall & Orig. State & 68.2 \\
                & AQM Recon & 66.7 \\\hline 
\end{tabular}
}
\caption{Results on RL probing tasks from \cite{anand2019unsupervised} with linear probe applied to original observation and to reconstructions from AQM after online compression. Acc is averaged for each game over game specific prediction. }\vspace{2pt}
\label{tab:rl}
\end{table}

\section{Conclusion}
We have introduced online continual compression. We demonstrated vector quantization can be used to control drift and how to create mechanisms that allow maintaining quality and maximizing memory usage. These allowed learning compression while compressing.
We have shown effectiveness of this online compression approach on standard continual classification benchmarks, as well as for compressing larger images, lidar, and atari data. We believe future work can consider dealing with temporal correlations for video and reinforcement learning tasks, as well as improved prioritization of samples for storage.

\nocite{langley00}

\bibliography{example_paper}
\bibliographystyle{icml2020}

\newpage
\appendix
\onecolumn


\section{Buffer Memory Management}\label{sec:mem}

We consider two settings to manage the samples inside the memory. In the first setting, we do not perform any codebook freezing. Therefore, as the model trains, the amount of compressed representations AQM can store increases smoothly. In this setting, the current amount of samples stored by AQM is a good approximation of the model's capacity. Therefore, we simply use this estimate instead of the total buffer size in the regular reservoir sampling scheme. The algorithm is presented in Alg \ref{algo:mem}.

\begin{algorithm2e}[H]\small
\SetAlgoLined
  \DontPrintSemicolon
    \KwIn{Memory $\mathcal{M}$ with capacity $C$ (bytes), sample $x$}
    
    $N_{reg} = \frac{C}{BYTES(x)}$
    
    
    capacity = $\max$(\ $N_{reg}$, NUM SAMPLES ($\mathcal{M}$)\ ) 
    
     \,\,\,\texttt{\%Probability of adding x} \\
    add $\sim \mathcal{B}(\frac{\text{capacity}}{\text{SAMPLE AMT SEEN SO FAR}})$ \texttt{\%Bernoulli}
    
    \,\,
    \If{add}
    {
        $hid_x$, $block_{id}$ = \textsc{Adaptive Compress}($x$, $AE$, $d_{th}$) \\
        \While {\text{BITS}($hid_x$) - \textsc{Free Space}($\mathcal{M}$) $> 0$}{
            \textsc{Delete Random}($\mathcal{M}$) \\
        }
        
    }
 \caption{AddToMemory}
 \label{algo:mem}
\end{algorithm2e}

This is not the case when we perform codebook freezing. In the latter setting, consider the moment when the first codebook is fixed; suddenly, the amount of samples the model can store has increased by a factor equal to the compression rate of the first block. Therefore, at this moment, the amount of samples currently stored by AQM is not a good approximation for the model's capacity. 

Moreover, when performing codebook freezing, since the capacity suddenly spikes, we must decide between a) having an imbalance in the buffer, where certain temporal regions of the streams are not equally represented, or not utilising all available memory and storing less incoming samples so they are in similar quantities as previous samples. We opt for the former approach, and propose a procedure that allows the buffer to rebalance itself as new training data becomes available. We illustrate the procedure with an example. 

\begin{figure}[h!]
    \centering
    \includegraphics[width=0.4\textwidth]{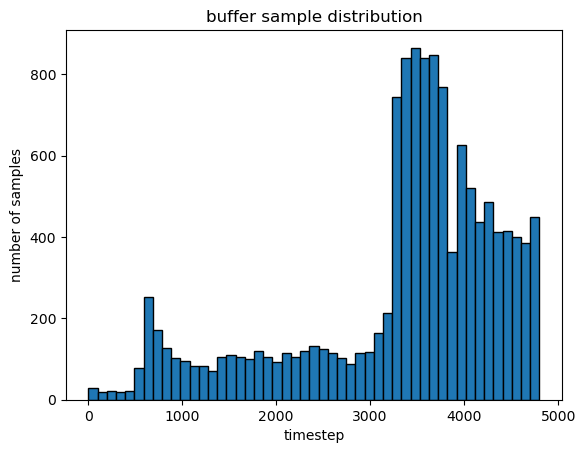}     \label{fig:buf}
    \includegraphics[width=0.4\textwidth]{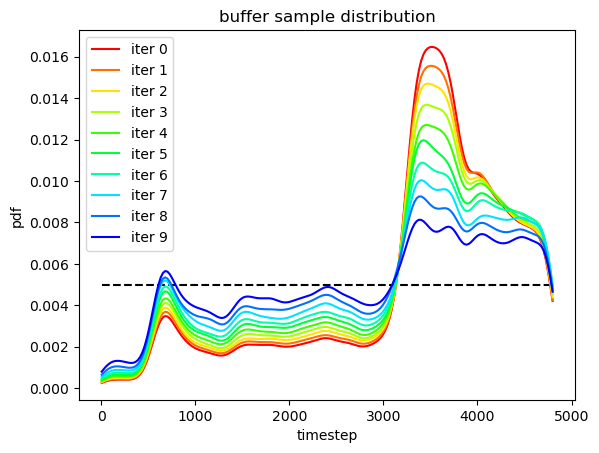}     \label{fig:kde}
    \caption{(left) histogram of samples in AQM where no buffer balancing was performed. (right) iterative buffer balancing procedure}
\end{figure}

Consider an AQM where the distribution of samples in the buffer is the one plotted in Fig \ref{fig:buf}. Specifically, we show the number of samples stored for each minibatch processed by the model. In this example, very few ($<$20) samples are stored from the earliest part of the stream, while a much larger number comes from the more recent part of the stream. Assuming that the model is over its memory capacity, we need to remove samples until the memory requirement is met. Ideally, we would like to remove more samples from parts of the stream where samples are abundant. In order to do so, we use Kernel Density Estimation on the histogram in Fig \ref{fig:buf}. Doing so gives us the line labelled $\texttt{iter0}$ in Fig \ref{fig:kde}. We then sample points according to the distribution given by $\texttt{iter0}$, remove them, and fit a new KDE with the remaining points (labelled $\texttt{iter1}$). In this example we repeat this procedure 10 times, until $\texttt{iter9}$. As we can see, the distribution of stored samples becomes closer to the uniform distribution, i.e. the setting where all parts of the streeam are equally represented. 

\vspace{3pt}

\vspace{10pt}
Therefore, in the setting where codebook freezing is performed, we first add all incoming points to the buffer. Then, points are removed according to the procedure described above. This allows for maximal memory usage while ensuring that the buffer self balances over time. Note that we need to store the timestamp alongside each sample, which has a negligible cost when dealing with high-dimensional inputs.

\section{Further Details of Experiments}

We include here further details regarding the models used in our experiments. For all reported results, \textbf{almost all hyperparameters are kept the same.} we set D the size of the embedding table equal to 100, we use a decay value of 0.6 for the Embedding EMA update, and the same architectural blocks. Across problems, we mainly vary the reconstruction threshold, as well how the blocks are stacked and their compression rates (by e.g. changing the number of codebooks per block).

\subsection{Cifar}
For CIFAR-10, we use a 1 block AQM,  latent size (16 x 16 x 100) is quantized  with (16 x 16 x 1) indices where the last index represents the number of codebooks. The codebook here contains 128 embeddings, giving a compression factor of 13.7$\times$. Due to the simplistic nature of the task and low resolution of the images, AQM already yields good compression before the end of the first task, hence adaptive compression and codebook freezing are not required. 
\newline
For the \citep{riemer2018learning} baseline, we ran a hyperparameter search to vary the compression size. Specifically, we ran a grid search for the number of categories per latent variable, as well as for the number of latent variables. We found the gumbel softmax much less stable during training and harder to cross-validate than vector quantization.

Below we show an example of the image quality of our approach compared to \cite{riemer2018learning}. We ran both AQM and  \citep{riemer2018learning} on the split CIFAR-10 task, then extracted images which happened to be in the buffer of both methods. 
\vspace{5pt}

\begin{figure}[h!]
    \centering
    \includegraphics[width=0.5\textwidth]{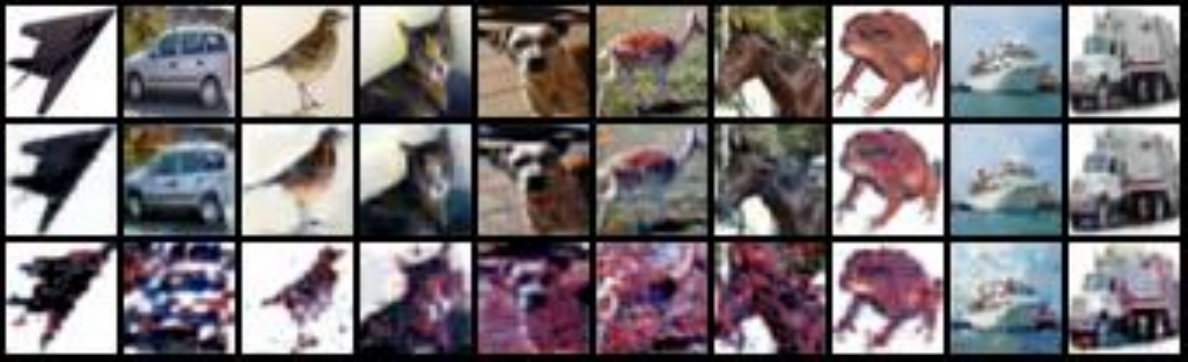}
    \caption{Bottom row: random buffer reconstructions using \cite{riemer2018learning}. Middle row: random buffer reconstructions using SQM. Top row: corresponding original image. Columns are ordered w.r.t their entry time in the buffer, from oldest to newest. All samples above were obtained from the disjoint CIFAR-10 task, and are 12$\times$ smaller than their original image.}
    \label{fig:dist_comparison}
\end{figure}

\subsection{Imagenet}

For the Offline 128 x 128 Imagenet experiment, we use the following three blocks to build AQM, with the following latent sizes : 
\begin{enumerate}
    \item (64 x 64 x 100) quantized using (64 x 64 x 1) indices, with a codebook of 16 vectors,  giving a 24$\times$ compression.
    \item (32 x 32 x 100) quantized using (32 x 32 x 1) indices, with a codebook of 256 vectors, giving a 48$\times$ compression.
    \item (32 x 32 x 100) quantized using (32 x 32 x 1) indices, with a codebook of 32 vectors,  giving a  76.8$\times$ compression.
\end{enumerate}

For the 2 block AQM, we searched over using blocks (1-2) (2-3) and (1-3). For 1 block AQM we simply tried all three blocks independently.

When stacking two blocks with the same latent sizes (e.g. block 2 and 3) the encoder and decoder functions for the second block are the identity. In other words, the second block simply learns another embedding matrix. 

\section{Drift Ablation}

Here we provide additional results for the drift ablation studied in $\ref{fig:ablate_2}$. We repeat the same experiment for different values of the reconstruction threshold parameter, which controls when the codebook freezing occurs. The table below shows that the same conclusion holds across multiple values for this parameter: codebook freezing yields little to no drift, while only negligibly hindering the model's ability to adapt to a distribution shift.

It is worth noting that in cases of severe drift (e.g. with $\texttt{recon th} = 7.5$) the model diverges because it is rehearsing on samples of poor quality. In this setting, codebook freezing performs better on both the streaming MSE and the drift MSE.

\vspace{10pt}

\begin{table*}[h!]
    \centering
    \resizebox{0.68\linewidth}{!}{%
    \begin{tabular}{|c|c|c|c||c|c||c}
        \texttt{recon th} & Freezing & Streaming MSE & Drift MSE & Streaming + Drift MSE  \\\hline \hline 
        1 &  No & 0.59 $\pm$ 0.03 & 1.01 $\pm$ 0.14 &   1.60 $\pm$ 0.17 \\ 
        
        1 & Yes & 0.60 $\pm$ 0.03 & 0.49 $\pm$ 0.02 &   \textbf{1.09 $\pm$ 0.05} \\ \hline
        
        2.5 &  No & 0.63 $\pm$ 0.06 & 13.24 $\pm$ 5.36  &   13.87 $\pm$ 5.42 \\ 
        
        2.5 & Yes & 0.81 $\pm$ 0.02 & 1.07 $\pm$ 0.05  &   \textbf{1.88 $\pm$ 1.12}  \\ \hline
        
        5 &  No & 0.65 $\pm$ 0.04 & 55.31 $\pm$ 36.82 &     55.96 $\pm$ 36.86\\ 
        
        5 & Yes & 0.92 $\pm$ 0.03 & 1.69 $\pm$ 0.18  & \textbf{2.61 $\pm$ 0.21} \\ \hline
        
        7.5 &  $\texttt{nan}$ & $\texttt{nan}$ & $\texttt{nan}$ &  $\texttt{nan}$ \\ 
        
        7.5 & Yes & 0.98 $\pm$ 0.11 & 2.10 $\pm$ 0.28   & \textbf{3.08 $\pm$ 0.39}\\ \hline

\end{tabular}} \label{tab:ablate2}
\caption{Here we provide additional results for the drift ablation discussion in section shown in Fig \ref{fig:ablate_2}. For clarity all results are multiplied by $100$. (e.g. $\texttt{recon th}$ of $2.5$ corresponds to $0.025$). Results averaged over 5 runs.}
\end{table*} 
\vspace{20pt}

\begin{figure}[H]
    \centering
    \includegraphics[width=0.45\textwidth]{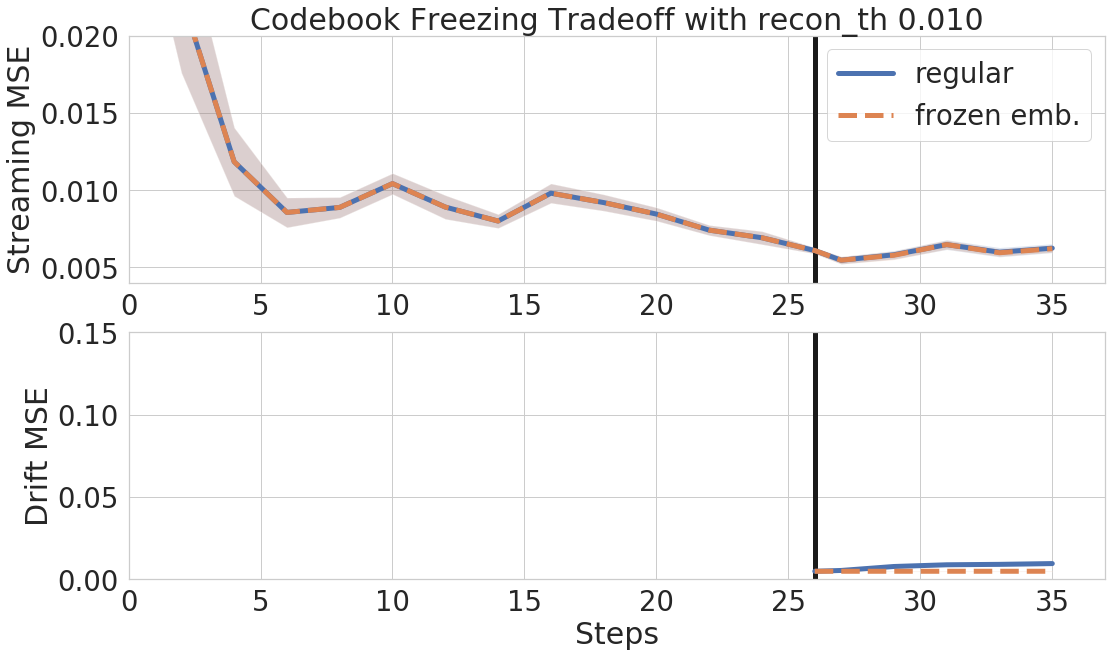} \vspace{15pt}
    \includegraphics[width=0.45\textwidth]{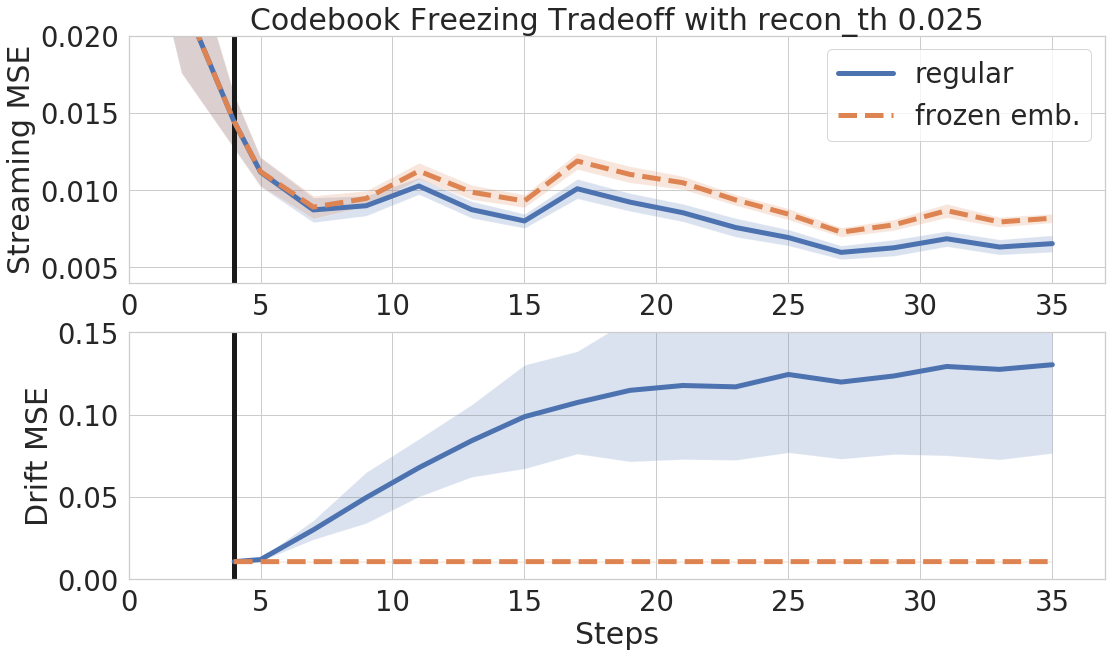} \\
    \includegraphics[width=0.45\textwidth]{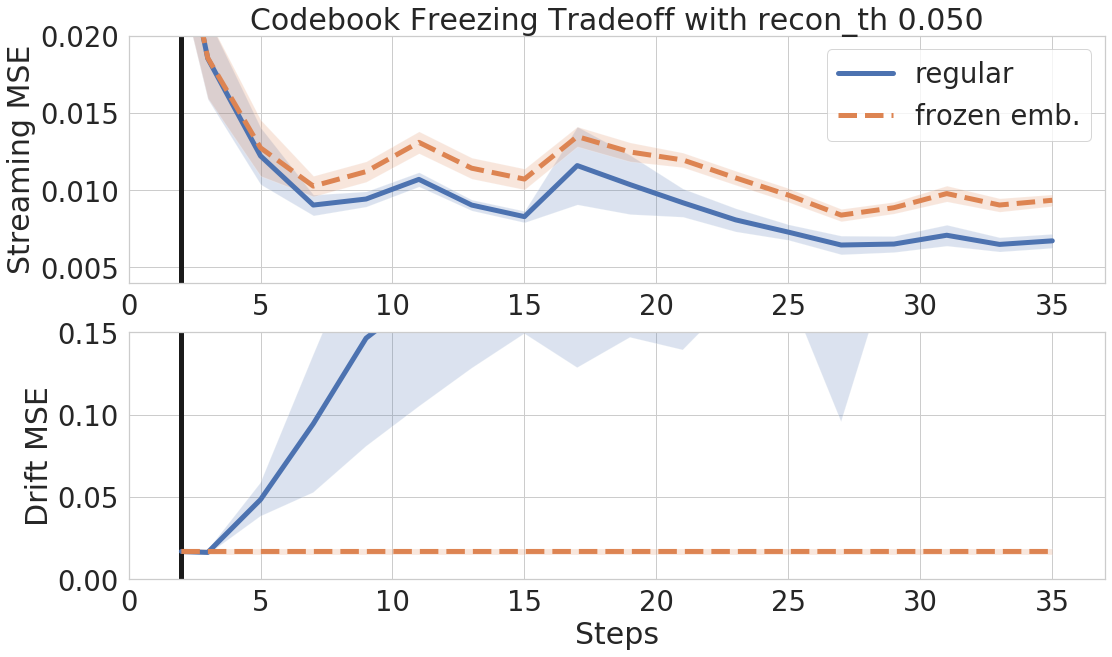} \vspace{15pt}
    \includegraphics[width=0.45\textwidth]{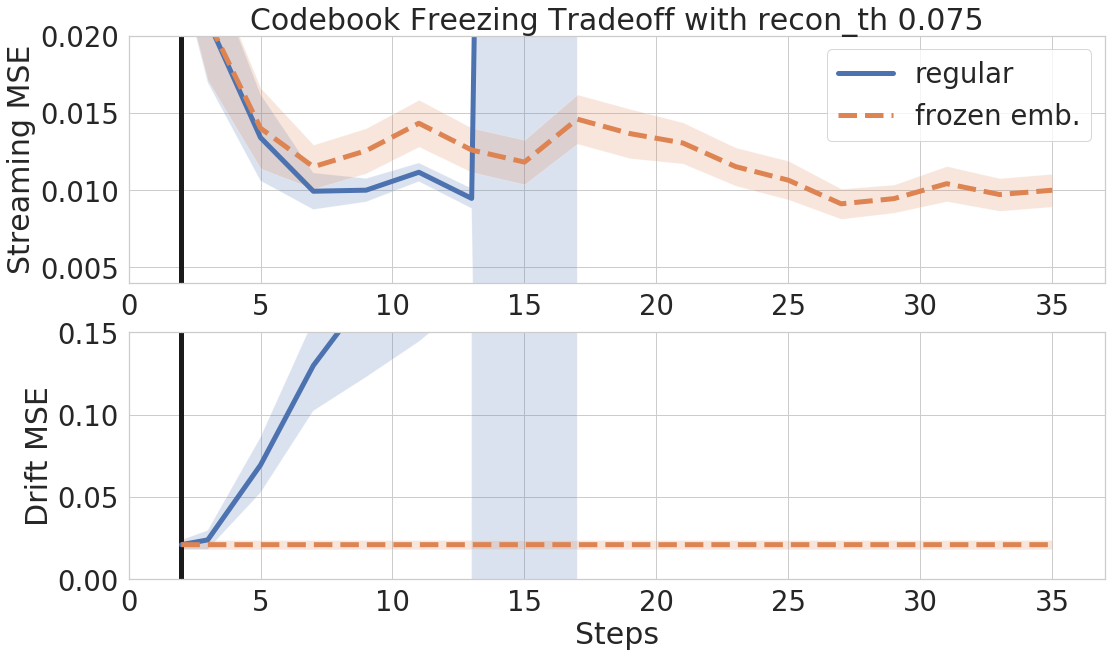} \\
    \caption{Visualization of results reported in Table \ref{tab:ablate2}. We kept the scale of the y axis consistent across the four graphs for easy comparison.}
\end{figure} 

\section{Atari}       

Below are sample reconstructions used in the Atari experiments. For each game, reconstructions are in the first row and the original uncompressed samples in the second row. Reconstructions are 16$\times$ smaller than the original RGB images. 

\begin{figure}[H]
    \centering
    \includegraphics[width=0.9\textwidth]{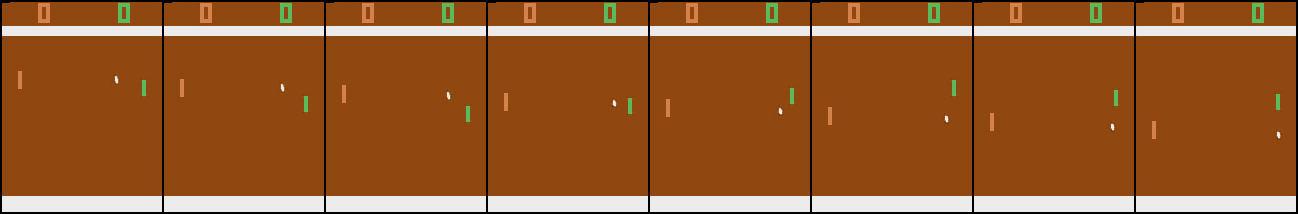} \vspace{15pt}
    \includegraphics[width=0.9\textwidth]{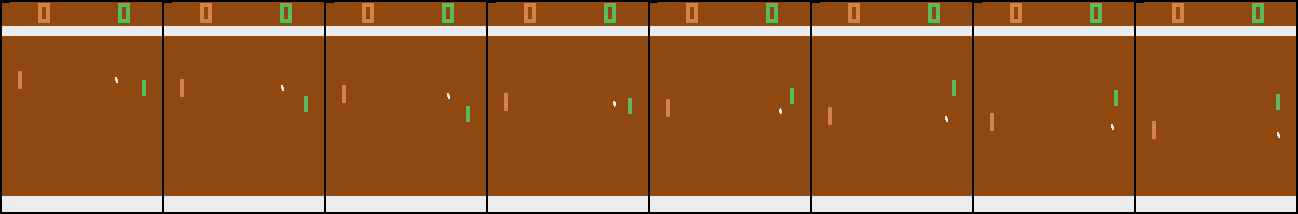} 
    \caption{Pong}
\end{figure}

\begin{figure}[H]
    \centering
    \includegraphics[width=0.9\textwidth]{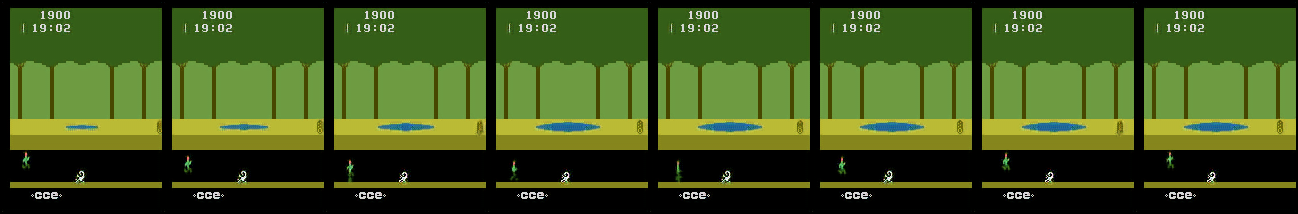} \vspace{15pt}
    \includegraphics[width=0.9\textwidth]{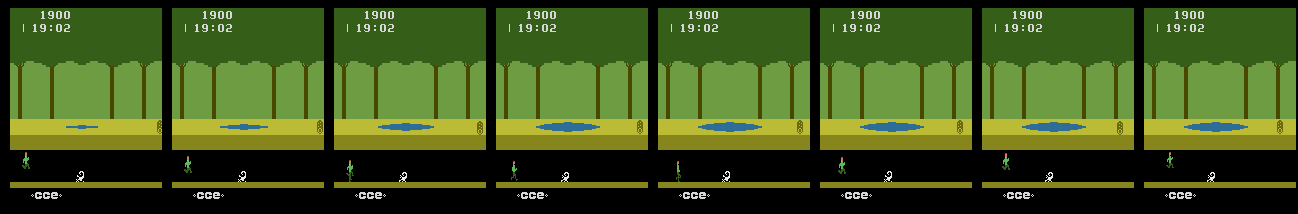}
    \caption{Pitfall}
\end{figure}

\begin{figure}[H]
    \centering
    \includegraphics[width=0.9\textwidth]{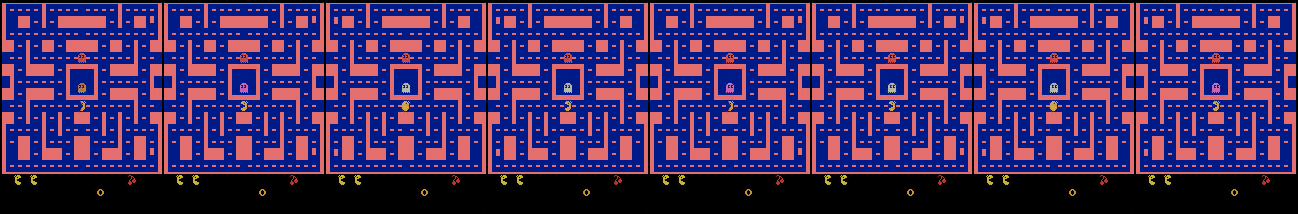}
    \includegraphics[width=0.9\textwidth]{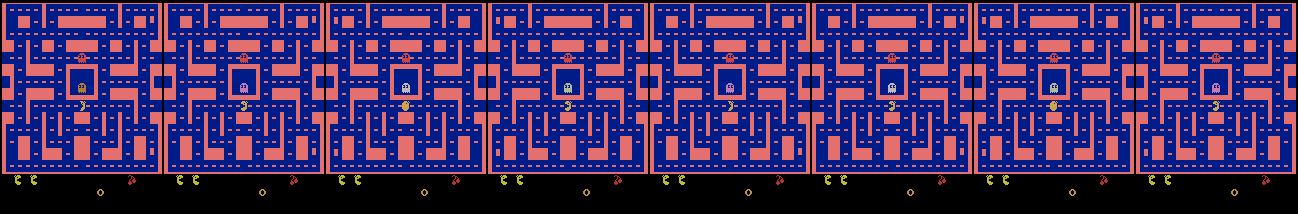}
    \caption{Ms Pacman}
\end{figure}

\section{Lidar Samples}
Here we show reveral lidar compressions (left) and their original counterpart (right). The compression rate is 32$\times$. We note that unlike RGB images, raw lidar scans are stored using floating points. We calculate the compression rate from the (smaller) polar projection instead of the 3 channel cartesian representation. 

\begin{figure}[H]
    \centering
    \includegraphics[width=0.45\textwidth]{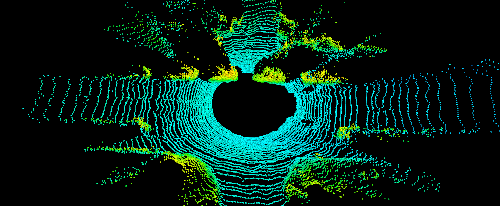} 
    \includegraphics[width=0.45\textwidth]{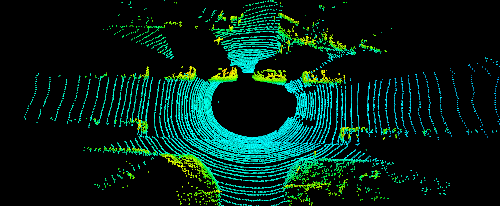} \\
    
    \includegraphics[width=0.45\textwidth]{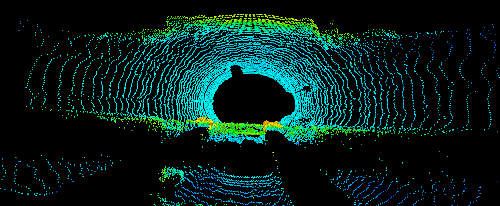} 
    \includegraphics[width=0.45\textwidth]{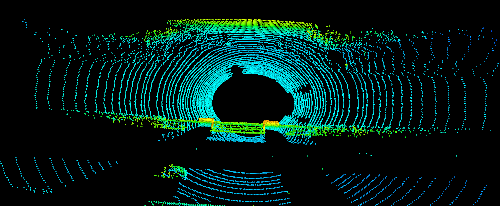} \\
    
    \includegraphics[width=0.45\textwidth]{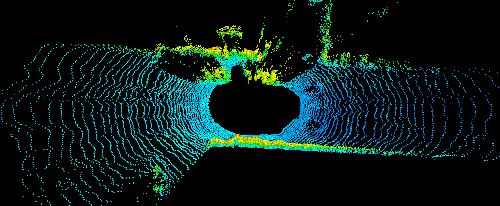} 
    \includegraphics[width=0.45\textwidth]{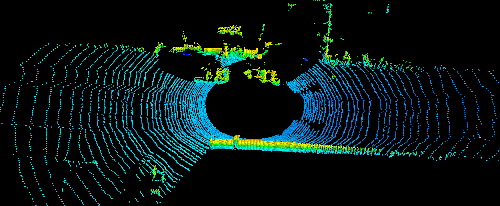} \\

    \includegraphics[width=0.45\textwidth]{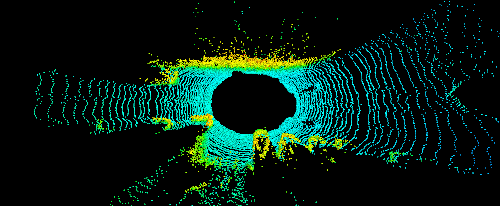} 
    \includegraphics[width=0.45\textwidth]{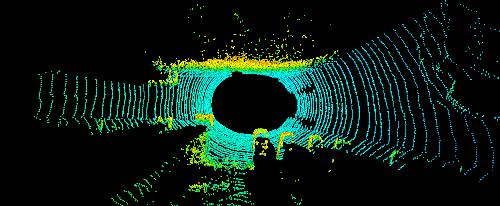} \\
    \caption{Lidar reconstruction (left) vs original (right)}
\end{figure}

\begin{figure}[H]
    \centering
    \includegraphics[width=0.45\textwidth]{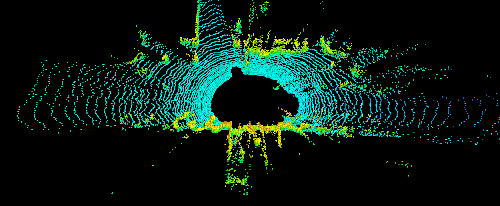} 
    \includegraphics[width=0.45\textwidth]{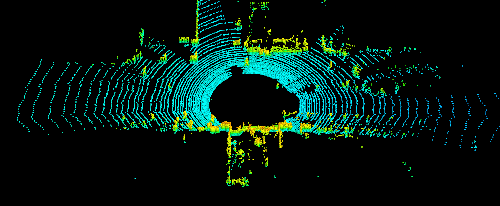} \\

    \includegraphics[width=0.45\textwidth]{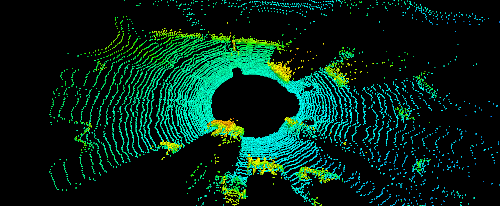} 
    \includegraphics[width=0.45\textwidth]{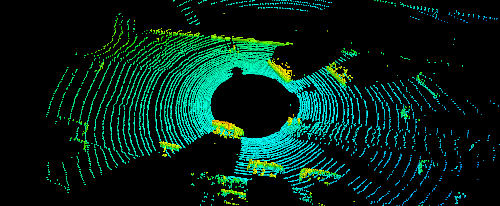} \\

    \includegraphics[width=0.45\textwidth]{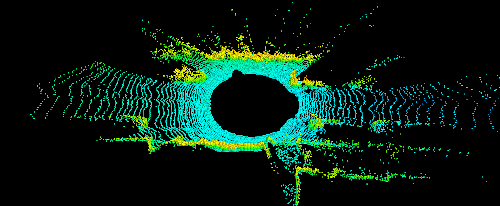} 
    \includegraphics[width=0.45\textwidth]{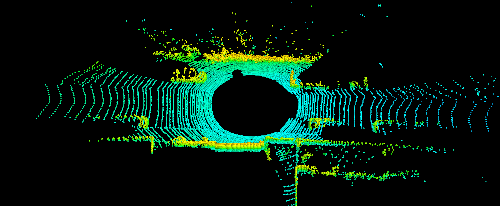} \\
    
    \includegraphics[width=0.45\textwidth]{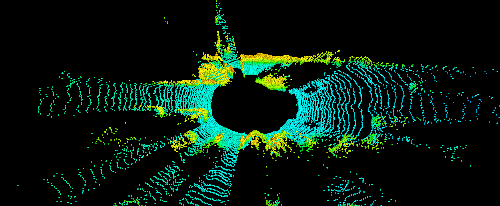} 
    \includegraphics[width=0.45\textwidth]{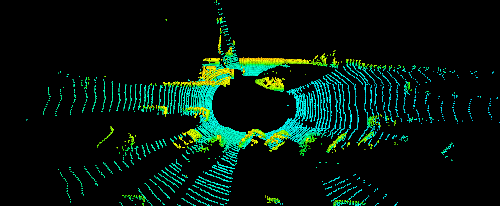} \\
    \caption{Lidar reconstruction (left) vs original (right)}
\end{figure}

\end{document}